\documentclass[sigconf]{acmart}
\pdfoutput=1
\usepackage{amsfonts}
\usepackage{booktabs} 
\usepackage{subfigure}
\usepackage{bm}
\usepackage{xspace}
\usepackage{multirow}
\usepackage{algorithm}
\usepackage{algpseudocode}

\usepackage{amssymb}
\usepackage[normalem]{ulem}
\usepackage{array}
\usepackage{graphicx}
\usepackage{multirow}
\usepackage{enumitem}

\usepackage{soul}
\usepackage{epstopdf}
\usepackage{setspace}
\usepackage{pifont}
\usepackage{xcolor}
\usepackage{mathrsfs}
\usepackage{amsmath}
\usepackage{ulem}

\newcommand{\paratitle}[1]{\vspace{1.5ex}\noindent\textbf{#1}}
\newcommand{\ie}{\emph{i.e.,}\xspace}

\newcommand{\eg}{\emph{e.g.,}\xspace}

\newcommand{\ignore}[1]{}

\newcommand{\tabincell}[2]{\begin{tabular}{@{}#1@{}}#2\end{tabular}}

\AtBeginDocument{%
	\providecommand\BibTeX{{%
			\normalfont B\kern-0.5em{\scshape i\kern-0.25em b}\kern-0.8em\TeX}}}

\copyrightyear{2020}
\acmYear{2020}
\setcopyright{acmcopyright}\acmConference[CIKM '20]{Proceedings of the 29th ACM International Conference on Information and Knowledge Management}{October 19--23, 2020}{Virtual Event, Ireland}
\acmBooktitle{Proceedings of the 29th ACM International Conference on Information and Knowledge Management (CIKM '20), October 19--23, 2020, Virtual Event, Ireland}
\acmPrice{15.00}
\acmDOI{10.1145/3340531.3411893}
\acmISBN{978-1-4503-6859-9/20/10}

\begin{document}
\fancyhead{}

\title{Knowledge-Enhanced Personalized Review Generation \\ with Capsule Graph Neural Network}

\author{Junyi Li$^{1,2}$, Siqing Li$^{3}$, Wayne Xin Zhao$^{1,2*}$}
\author{Gaole He$^{3}$, Zhicheng Wei$^{4}$, Nicholas Jing Yuan$^{4}$ and Ji-Rong Wen$^{1,2}$}
\thanks{$^*$Corresponding author.}
\affiliation{%
	\institution{$^1$Gaoling School of Artificial Intelligence, Renmin University of China}
	\institution{$^2$Beijing Key Laboratory of Big Data Management and Analysis Methods}
	\institution{$^3$School of Information, Renmin University of China}
	\institution{$^4$Huawei Cloud \& AI}
}
\affiliation{%
	\institution{\{lijunyi,lisiqing,hegaole,jrwen\}@ruc.edu.cn, batmanfly@gmail.com, \{weizhicheng1,nicholas.yuan\}@huawei.com}
}

\begin{abstract}
Personalized review generation (PRG) aims to automatically produce review text reflecting user preference, which is a challenging natural language generation task. Most of previous studies do not explicitly model  factual description of products, tending to generate uninformative content. Moreover, they mainly focus on word-level generation, but cannot accurately reflect more abstractive  user preference in multiple aspects.

To address the above issues, we propose a novel knowledge-enhanced PRG model 
based on capsule graph neural network~(Caps-GNN). We first 
construct a heterogeneous knowledge graph (HKG) for utilizing rich item attributes.
We adopt  Caps-GNN to learn graph capsules for encoding underlying characteristics from the HKG. Our generation process contains two major steps, namely aspect sequence generation and sentence generation. First, based on graph capsules, we adaptively learn aspect capsules for inferring the aspect sequence.   Then, conditioned on the inferred aspect label, we design a graph-based copy mechanism to generate sentences by incorporating related entities or words from HKG.
To our knowledge, we are the first to utilize knowledge graph for the PRG task.
The incorporated KG information is able to enhance user preference at both aspect and word levels. Extensive experiments on three real-world datasets have demonstrated the effectiveness of our model on the PRG task.
\end{abstract}




\maketitle

\section{Introduction}

With the rapid development of e-commerce, online reviews written by  users  have become increasingly important for reflecting real customer experiences. To ease the process of review writing, the task of personalized review generation~(PRG)~\cite{ZhouYHZXZ18,LiT19} has been proposed to automatically produce review text conditioned on necessary context data, \eg users, items, and ratings.

As a mainstream solution, RNN-based models  have been widely applied to the PRG task~\cite{ZhouYHZXZ18,ZangW17}. Standard RNN models mainly model sequential dependency among tokens,  which cannot effectively generate high-quality review text.
Many efforts have been devoted to improving this kind of architecture for the PRG task, including context utilization~\cite{ZhouYHZXZ18},  long text generation~\cite{LiZWS19}, and  writing style enrichment~\cite{LiT19}.
These studies have improved the performance of the PRG task to some extent. However, two major issues still remain to be solved.
First, the generated text is likely  to be uninformative, lacking factual description on product information. Although several studies try to incorporate structural or semantic features (\eg aspect words \cite{NiM18} and history corpus \cite{LiT19}),  they mainly extract such features from the review text.  
Using review data alone, it is difficult to fully capture diverse and comprehensive facts from unstructured text.
Second, most of these studies focus on word-level generation, which makes it difficult to directly model  user preference at a higher level. For example, given a product, a user may focus on the \emph{price}, while another user may emphasize the \emph{look}.

To address these issues, we propose to improve the PRG task with external knowledge graph (KG).
By associating online items
with KG entities~\cite{zhao2019kb4rec,HuangZDWC18}, we are able to obtain rich attribute or feature information for items, which is potentially useful for the PRG task.
Although the idea is intuitive, it is not easy to fully utilize the knowledge information for generating review text in our task.
KG typically organizes facts as triples, describing the relation between two involved entities.
It may not be suitable to simply integrate KG information to enhance text representations or capture user preference due to varying intrinsic characteristics of different data signals.

\begin{figure}[tb]
	\centering
	\includegraphics[width=0.45\textwidth]{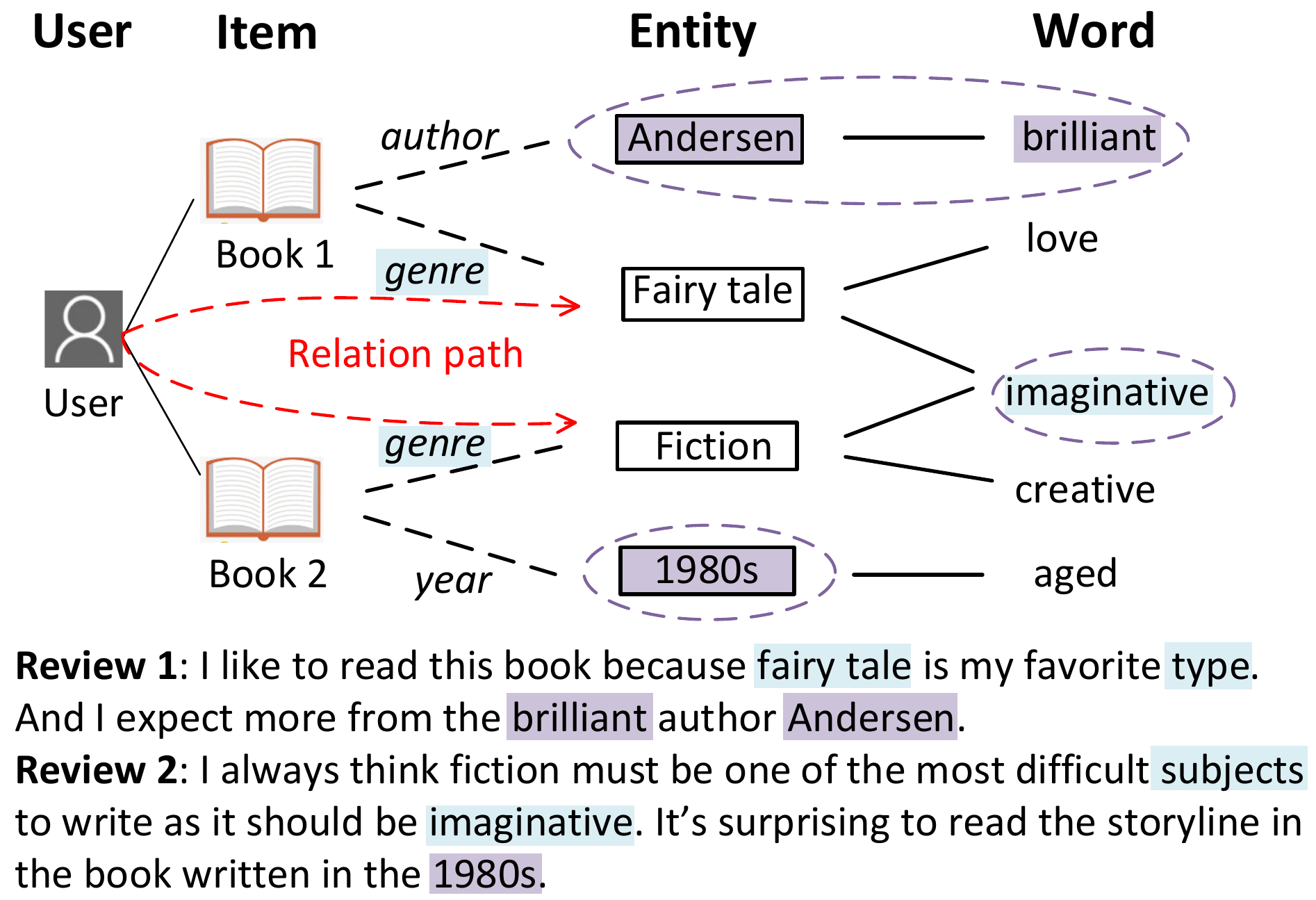}
	\caption{An illustrative heterogeneous knowledge graph~(HKG) example on \textsc{Amazon} Book dataset. It captures the user preference at both  aspect and word levels.}
	\label{fig-example}
\end{figure}

In order to bridge the semantic gap, we augment the original KG with user and word nodes, and construct a heterogeneous knowledge graph (HKG) by adding user-item links and entity-word links.
User-item links are formed according to user-item interactions, and entity-word links are formed according to their co-occurrence in review sentences.
We seek to learn a unified semantic space that is able to encode different kinds of nodes.
Figure~\ref{fig-example} presents an illustrative example for the HKG.
Given such a graph, we focus on two kinds of useful information for the PRG task.
First, the associated facts regarding to an item (\eg the author of a book is \emph{Andersen}) can be incorporated to enrich the review content. Second, considering users as target nodes, we can utilize this graph to infer users' preference  on some specific relation or aspect (\eg \emph{genre} or \emph{subject}). The two kinds of information reflect word- and aspect-level enrichment, respectively. To utilize the semantics at the two levels, we decompose 
the review generation process into two stages, namely aspect sequence generation and sentence generation.  We aim to inject multi-granularity KG information in different generation stages for improving the PRG task. 

To this end, in this paper, we propose a KG-enhanced personalized review generation model based on capsule graph neural networks~(Caps-GNN).
Compared with most of existing GNN-based methods representing graphs as individual scalar features~\cite{HamiltonYL17,ZhangCNC18}, Caps-GNN can extract underlying characteristics of graphs as \emph{capsules} at the graph level through the dynamic routing mechanism and each capsule reflects the graph properties in different aspects.
Based on the constructed HKG, we utilize Caps-GNN to extract graph properties in different aspects as \emph{graph capsules}, which may be helpful to infer aspect- and word-level user preference.
For aspect sequence generation, we propose a novel adaptive learning algorithm that is able to capture personalized user preference at the aspect level, called \emph{aspect capsules}, from the graph capsules. 
We associate an aspect capsule with a unique aspect from unsupervised topic models.  
Furthermore, for the generation of sentences, we utilize the learned aspect capsules to capture personalized user preference at the word level.
Specially, we design a graph-based copy mechanism to generate related entities or words by copying them from the HKG, which can enrich the review contents. 
In this way, KG information has been effectively utilized  at both aspect and word levels in our model.

To our knowledge, we are the first to utilize KG to capture both aspect- and word-level user preference for generating personalized review text.
For evaluation, we constructed three review datasets by associating items with KG entities.
Extensive experiments  demonstrate the effectiveness of KG information and our model. 

\section{Related Work}
Recently, many researchers have made great efforts on the natural language generation (NLG) task~\cite{ChengXGLZF18, ZhaoZWYHZ18, LewisDF18}. Automatic review generation is a specific task of NLG, which focuses on helping online users to generate product reviews~\cite{LiZWS19,ZhouYHZXZ18}.

Typical methods adopted RNNs to model the generation process and utilize available context information, such as user, item and rating~\cite{ZangW17,ZhouYHZXZ18}.  In order to avoid repetition issue caused by the RNN models and generate long and diverse texts, Generative Adversarial Nets (GAN) based approaches have been applied to text generation~\cite{GuoLCZYW18,YuZWY17}.
However, the generation process is unaware of  the underlying semantic structure of text. 
To make the generated text more informative, several studies utilized side information with a more instructive  generation process~\cite{NiM18,LiZWS19,LiT19}. These works utilize context features, \eg aspect words~\cite{NiM18} and history corpus~\cite{LiT19}, to enrich the generated content. 
While, their side information was mainly mined from the review itself, which cannot fully cover diverse and rich semantic information.
We are also aware of the works that utilize structural  knowledge data to enrich the diversity of generated texts~\cite{SunDZMSC18}. 
However, these studies do not utilize knowledge information to learn the writing preference of users.

Furthermore, closely related to the recommendation task, several studies attempted to model the interactions between user and product with review as explanation~\cite{NiLM19,CatherineC17}. They mainly capture the adoption preference over items, while, we focus on the writing preference for review generation. They still rely on the review text itself for learning useful explanation for users' adoption behaviors.  
The focus of this work is to explore external KG data for extracting effective information for the PRG task. 

Our work is inspired by the work of capsule graph neural network~\cite{XinyiC19}, especially its application on aspect extraction~\cite{du-etal-2019-capsule, ChenQ19}. These works mainly focus on capsule networks for aspect-level sentiment classification. 
While, our work focuses on inferring aspect information using KG data for review generation.

\section{Problem Formulation}
\label{preliminary}

In this section, we introduce the notations that will be used throughout the paper, and then formally define the task.

\paratitle{Basic Notations}.
Let $\mathcal{U}$ and $\mathcal{I}$ denote a user set and an item set, respectively. 
A review text is written by a user $u \in \mathcal{U}$ about an item $i \in \mathcal{I}$ with the content on some specific \emph{aspects}. Here, we introduce the term of  ``\emph{aspect}'' to describe some properties about an item (\eg price and service for a restaurant). Following~\cite{BrodyE10,JoO11,TitovM08}, we assume that a sentence (or a shorter text segment) is associated with a single aspect label, and  aspect labels can be obtained in some unsupervised way (\eg topic models~\cite{ZhaoJWHLYL11}).
Formally, a review text  is denoted by $w^{1:m}=\{\langle w_{j,1},\cdots,w_{j,t},\cdots,w_{j,n_j} \rangle \}_{j=1}^m$, consisting of $m$ sentences, where $w_{j,t}$ denotes the $t$-th word (from a vocabulary  $\mathcal{V}$) of the $j$-th review sentence and $n_j$ is the length of the $j$-th sentence.
 Let $\mathcal{A}$ denote a set of $A$ aspects in our collection. The aspect sequence of a review text is denoted by $a^{1:m}={\langle a_1,\cdots,a_j,\cdots,a_m \rangle}$, where $a_j \in \mathcal{A}$ is the aspect label of the $j$-th sentence.

\paratitle{Aligning Items to Knowledge Graph Entities}. In our task, 
a knowledge graph~(KG) $\mathcal{T}$ is given as input. 
Typically, KG stores the  information in fact triples:  $\mathcal{T}=\{\langle h,r, t \rangle\}$, where each triple
describes that there is a relation $r$ between head entity $h$ and tail entity $t$ regarding to some facts.
Furthermore, we assume that an  item  can be aligned to a KG entity.
For instance, the Freebase movie entity 
``\emph{Avatar}'' (with the Freebase ID \emph{m.0bth54}) has an entry of a movie item in IMDb (with the IMDb ID \emph{tt0499549}).
Several studies~\cite{zhao2019kb4rec,HuangZDWC18} try to develop heuristic algorithms for  \emph{item-to-entity alignment} and have released public linkage dataset. It is easier to obtain such a data alignment in some specific application when there is a domain-specific KG constructed by the enterprise. 


\paratitle{Heterogeneous Knowledge Graph}. 
In order to better utilize KG information for our task, we  introduce a \emph{heterogeneous knowledge graph}~(HKG) $\mathcal{G}$ for extending original KG by adding user and word nodes.
We create user-item links according to their interaction relations (\ie review writing), and create entity-word links according to their  co-occurrence relation in the review sentences. 
In this way, the HKG $\mathcal{G}$ can be written as: $\mathcal{G}=\mathcal{T}\cup\{ \langle u, r_{int}, i \rangle\} \cup\{ \langle  e, r_{co}, w \rangle\}$, where $r_{int}$ and $r_{co}$ denote the relations of user-item interaction and entity-word  co-occurrence, respectively.
Figure~\ref{fig-example} presents an illustrative example for our HKG. 
Such a KG is useful to infer users' preference about item properties via some \emph{user-to-entity} relation paths, and capture semantic relatedness between entities and words via \emph{entity-word} links. 
For example, in Fig.~\ref{fig-example}, the user prefers to comment on the \emph{genre} relation, and the entity \emph{Andersen} is  associated with the modifier word \emph{brilliant}. 
Such an example indicates that KG data is likely to be useful in review generation by providing relation or entity related information. 

\paratitle{Task Definition}. Personalized review generation (PRG)~\cite{NiM18,ZhouYHZXZ18} aims to automatically produce the review text for  user $u$ on item $i$ given his/her rating score $s$ and possible context information if any. We follow \cite{LiZWS19} to consider an aspect-aware generation process:  an aspect sequence is first generated and then a sentence conditioned on an aspect label is subsequently generated.
In our setting, the task of PRG can be formulated to seek a model (parameterized by $\Theta$) by maximizing the joint probability of the aspects and word sequences through the training collection: 
\begin{equation}\label{eq-joint-prob}
\sum_{\langle w^{1:m},  a^{1:m} \rangle } \log \text{Pr}(w^{1:m},  a^{1:m} |c , \mathcal{G}; \Theta),
\end{equation}
where we have  the context information $c=\{u,i,s\}$ and the HKG $\mathcal{G}$ as  input. Different from previous works~\cite{ZhouYHZXZ18,NiM18}, we construct and incorporate the HKG $\mathcal{G}$ as available resource for review generation. We would like to utilize KG information in the above two generation stages, capturing both aspect- and word-level user preference.

\begin{figure}[!tb]
	\centering 
	\includegraphics[width=0.45\textwidth]{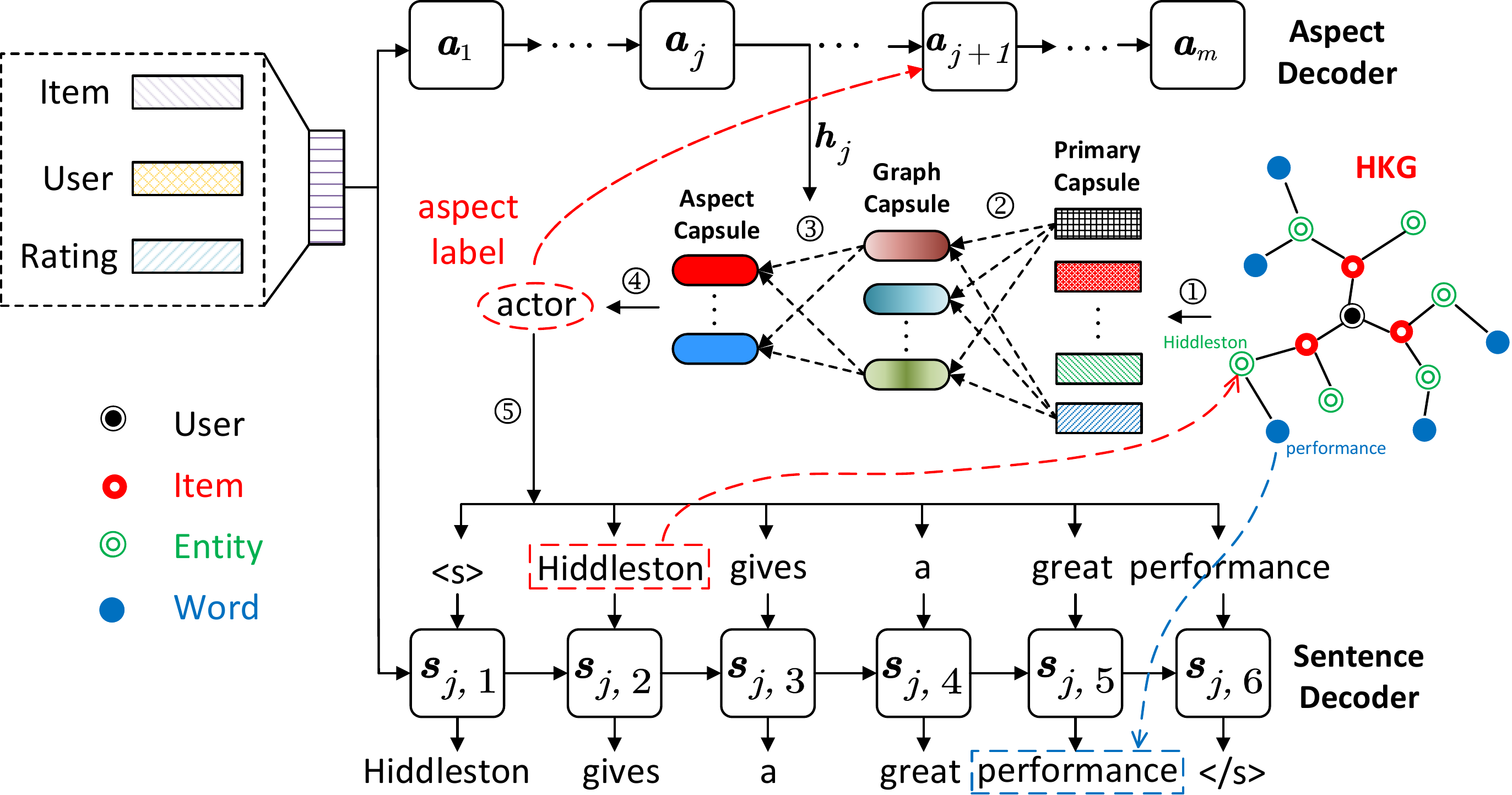} 
	\caption{The overview of the proposed generative process with the example of ``Hiddleston gives a great performance''. The predicted aspect label is ``\emph{actor}'', the previous token ``\emph{Hiddleston}'' is used as a query to find a neighboring word node, and the word node ``\emph{performance}'' is selected. } 
	\label{fig-model}
\end{figure}

\section{The Proposed Approach}
In this section, we present the proposed KG-enhanced  review generation model. 
We first introduce a capsule graph  neural network for learning graph capsules encoding the graph characteristics in different aspects. By utilizing the learned graph capsules, we further design KG-enhanced models for both aspect generation and sentence generation. 
Figure~\ref{fig-model} presents  an overview illustration of the proposed model. Next we will describe each part in detail.

\subsection{Graph Capsule Learning} 
The major purpose of graph capsule learning is to encode HKG information for capturing user preference in different aspects. 
For this purpose, 
we propose to use Capsule Graph Neural Network (Caps-GNN)~\cite{XinyiC19} to generate high-quality graph embeddings for the HKG, called \emph{graph capsules}. Graph capsules reflect the properties of the HKG in different aspects.
We use $Z$  graph capsules to encode the graph, denoted by $\bm{P} \in \mathbb{R}^{Z \times d_C}$, where $d_C$ is the embedding size of a graph capsule.  
Each graph capsule encodes the characteristics of the graph related to some specific dimension or property.
Graph capsules are derived based on primary  capsules via dynamic routing. We first describe how to learn primary  capsules.

\subsubsection{Learning Primary Capsules } For convenience, 
we use a general placeholder $n$ ($n_j$ and $n_k$) to denote any node on HKG $\mathcal{G}$.  Let $\bm{v}_n \in \mathbb{R}^{d_E}$  denote the node embedding for a general node $n$, where $d_E$ is the embedding size. Node embeddings  can be initialized with pre-trained KG embeddings or word embeddings~\cite{MikolovSCCD13, YangYHGD14a}. 
We use R-GCN~\cite{SchlichtkrullKB18} to extract node embeddings from different layers. The embedding of node $n_j$ in $(l+1)$-th layer  can be computed via: 

\begin{equation}\label{eq-RGCN}
\bm{v}^{(l+1)}_{n_j}= \sigma (\sum_{r \in \mathcal{R}} \sum_{n_k \in \mathcal{N}^r_{n_j}} \bm{W}^{(l)}_r \bm{v}^{(l)}_{n_k} + \bm{W}^{(l)}_0 \bm{v}^{(l)}_{n_j} ),
\end{equation}
where $W^{(l)}_r$ and $W^{(l)}_0$ are the trainable matrices, and $\mathcal{N}^r_{n_j}$ denotes the set of neighbors of node $n_j$ under relation $r$ from the relation set $\mathcal{R}$. 
After  stacking the R-GCN layer by $L$ times, we concatenate the embeddings of a node $n_j$ over the $L$ layers into a vector, denoted by $\bm{x}_j \in \mathbb{R}^{L\cdot d_E}$, 
which represents the $j$-th  \emph{primary capsule}.


\subsubsection{Dynamic Routing  for Graph Capsule} With primary capsules, following~\cite{XinyiC19}, dynamic routing mechanism is applied to generate graph capsules $\bm{P} \in \mathbb{R}^{Z \times d_C}$, where $Z$ is the number of graph capsules and $d_C$ is the dimension of graph capsule. Each graph capsule $\bm{p}_z \in \mathbb{R}^{d_C}$ is computed via a non-linear ``squashing'' function:
\begin{equation}\label{eq-squashing}
\bm{p}_z = \frac{\left\|\bm{s}_z\right\|_2}{1+\left\|\bm{s}_z\right\|_2} \frac{\bm{s}_z}{\left\|\bm{s}_z\right\|_2}, 
\end{equation}
where $\bm{p}_z$ is the $z$-th graph capsule and $\bm{s}_z$ is its total input.
The total input $\bm{s}_z$ is a weighted sum over all ``prediction vector'' $\bm{\hat{x}}_{z|j}$, which is produced by multiplying the primary capsule $\bm{x}_j$ with a weight matrix $\bm{W}_{jz}$:
\begin{eqnarray}
\bm{s}_z &=& \sum_j c_{jz} \bm{\hat{x}}_{z|j}, \\
\bm{\hat{x}}_{z|j} &=& \bm{W}_{jz}\bm{x}_j,  \nonumber
\end{eqnarray}
where $c_{jz}$ are coupling coefficients indicating the importance of primary capsule $\bm{x}_j$ with respect to graph capsule $\bm{p}_z$. The coupling coefficients are determined by a ``routing softmax'':
\begin{equation}
c_{jz} = \frac{\text{exp}(b_{jz})}{\sum_j \text{exp}(b_{jz})}.
\end{equation}
The initial logits $b_{jz}$ are the log prior probabilities. We employ the dynamic routing mechanism for multiple iterations, and the logits can be iteratively updated as follows:
\begin{equation}\label{eq-logits}
b_{jz} = b_{jz} + \bm{\hat{x}}_{z|j}^\top \bm{p}_z.
\end{equation}

\subsection{Capsule-based Aspect Generation}
\label{graph-based-aspect}
We develop the aspect generation module based on an encoder-decoder framework.
We assume that aspect labels of sentences are provided as input for training. 
Our main idea is to infer personalized user preference over item aspects based on the HKG. 

\subsubsection{Basic Aspect Decoder} 

We adopt the GRU-based RNN network using graph capsules $\bm{P}$ to develop the aspect decoder. Let $\bm{h}_j \in \mathbb{R}^{d_{H}}$ denote a $d_{H}$-dimensional hidden vector at the $j$-th time step, which is computed via:
\begin{equation}\label{eq-aspect-gru}
\bm{h}_j = \text{GRU}(\bm{h}_{j-1}, \bm{v}_{a_{j-1}}),
\end{equation}
where $\bm{v}_{a_{j-1}} \in \mathbb{R}^{d_A}$ is the embedding of the previous aspect label $a_{j-1}$. Following~\cite{ZhouYHZXZ18}, the hidden vector of the first time step can be initialized with the context embedding $ \bm{v}_c$:
\begin{equation}\label{eq-MLP}
\bm{h}_0 \leftarrow \bm{v}_c= \text{MLP}([\bm{v}_u; \bm{v}_i; \bm{v}_s]).
\end{equation}

\subsubsection{Learning Adaptive Aspect Capsules} 
At the $j$-th time step, we can obtain the hidden state vector $\bm{h}_j$ from the previous aspect sequence. We further utilize $\bm{h}_j$ as a ``query'' to ``read'' important parts (denoted by adaptive graph capsules $\widetilde{\bm{P}}$) from the graph capsules by using an attention mechanism~\cite{LuongPM15}:
 \begin{equation}\label{eq-rescale-cap}
\tilde{\bm{p}}_z = \frac{\exp(\tanh(\bm{W}_1 [\bm{p}_z; \bm{h}_j]))}{\sum_{z'=1}^{Z} \exp(\tanh(\bm{W}_1 [\bm{p}_{z'}; \bm{h}_j]))} \bm{p}_z,
\end{equation}
where  $\bm{W}_1$ is a parameter matrix, and $\tilde{\bm{p}}_z \in \widetilde{\bm{P}}$ is the $z$-th adaptive graph capsule. In this way, our model can generate personalized aspect sequence by adaptively focusing on different parts of the HKG in each time step. 
Finally, dynamic routing mechanism (see Eq. \ref{eq-squashing}-\ref{eq-logits}) is applied again over the adaptive graph capsules $\widetilde{\bm{P}}$ to generate final aspect capsules $\bm{Q} \in \mathbb{R}^{ A \times d_C}$, where $A$ is the number of aspect labels and $d_C$ is the dimension of aspect capsule. The length of capsules reflects the probability of the presence of aspects at the current time step. Finally, the $j$-th aspect label $a_j$ is predicted via:
\begin{equation}\label{eq-aspect-prob}
a_j =  \arg\max_k \left\|\bm{q}_k\right\|_2,
\end{equation}
where $\bm{q}_k \in \mathbb{R}^{d_C}$ is the $k$-th aspect capsule.
 
To learn the aspect capsules, we adopt a margin based loss for $j$-th time step:
\begin{equation}\label{eq-aspect-loss}
L_j = \max(0, m^+ - \left\|\bm{q}_{a_j}\right\|)^2 + \lambda \sum_{i \neq a_j}\max(0, \left\|\bm{q}_i\right\| - m^-)^2, 
\end{equation}
where  $m^+=0.9$, $m^-=0.1$ and $\lambda=0.5$ following \cite{SabourFH17}.

\subsection{KG-enhanced Sentence Generation}
Given the inferred aspect labels, we study how to generate the text content of a sentence.  
We  start with a base sentence decoder by using GRU-based network, and then extend it by incorporating KG-based copy mechanism.

\subsubsection{Base Sentence Decoder}

The base sentence generation module adopts a standard attentional encoder-decoder architecture. 
Intuitively,  the descriptive words for different aspects are likely to be varying. Hence, we need to consider the effect of aspect labels for word generation. 
Let $ \bm{s}_{j,t} \in \mathbb{R}^{d_S}$ denotes the $d_S$-dimensional hidden vector at the $t$-th time step for the $j$-th sentence, which is computed via:
\begin{equation}\label{eq-sentence-gru}
\bm{s}_{j,t} = \text{GRU}(\bm{s}_{j,t-1}, \bm{x}_{j,t}),
\end{equation} 
where $\bm{x}_{j,t}$ is further defined as the element-wise product between the embedding of the previous sentence word $\bm{v}_{w_{j,t-1}} \in \mathbb{R}^{d_W}$ and the embedding of the current aspect capsule $\bm{q}_{a_j}$:
\begin{equation}\label{eq-sentence-input}
\bm{x}_{j,t}= \bm{v}_{w_{j,t-1}}\odot \bm{q}_{a_j}.
\end{equation}
In this way, the adaptive aspect information can be utilized at each time step to generate a personalized word sequence.
Following~\cite{ZhouYHZXZ18}, we also apply standard attention mechanisms to attend to both context information and previous tokens 
for improving the state representation, and obtain a context vector $\tilde{\bm{c}}_{j,t}$.
With $\tilde{\bm{c}}_{j,t}$, we can generate a word according to a softmax probability function:
\begin{eqnarray}\label{eq-att-gru}
\text{Pr}_{1}(w|a_j, c) &=& \text{softmax}(\bm{W}_3 \tilde{\bm{s}}_{j,t} + \bm{b}_1), \\
\tilde{\bm{s}}_{j,t} &=& \tanh (\bm{W}_2 [\tilde{\bm{c}}_{j,t}; \bm{s}_{j,t}]). \nonumber
\end{eqnarray}

\subsubsection{Incorporating KG-based Copy Mechanism}  

As shown in Fig.~\ref{fig-example}, we organize words and entities as nodes on the HKG. Inspired by models for the question-answering tasks~\cite{SunDZMSC18}, our decoder attentively reads the history and context information to form queries, then adaptively chooses a personalized word or entity from the HKG for sentence generation. Such a way can be effectively modeled using the copy mechanism. 
We assume that the predictive probability  of a word can be decomposed into two parts, either generating a word or copying a node from the HKG:
\begin{eqnarray}\label{eq-sentence-prob}
&&\text{Pr}(w_{j,t}=w|w_{j,<t}, a_j, c, \mathcal{G}) \\
&=& \alpha \cdot \text{Pr}_{1}(w|a_j, c) + (1-\alpha) \cdot \text{Pr}_{2}(w|a_j, c, \mathcal{G}),\nonumber
\end{eqnarray} 
where $\text{Pr}_{1}(w|a_j, c)$ is the generative probability from our base decoder defined in Eq.~\ref{eq-att-gru}, and 
$\text{Pr}_{2}(w|a_j, c, \mathcal{G})$ is the copy probability of a word defined as below:
\begin{eqnarray}
\text{Pr}_{2}(w|a_j, c, \mathcal{G})&=& \text{softmax} (\bm{W}_5 \check{\bm{s}}_{j,t} + \bm{b}_2),\\ 
\check{\bm{s}}_{j,t} &=& \tanh(\bm{W}_4 [\tilde{\bm{c}}_{j,t}; \bm{s}_{j,t}; \bm{v}_w]),\nonumber
\end{eqnarray}
\noindent where $\bm{v}_w$ is the embedding of an entity or a word node $w$ learned with Caps-GNN. Considering the efficiency, we only enumerate the nodes that are at least linked to a previous token in the generated sub-sequence.  
In Eq.~\ref{eq-sentence-prob}, we  dynamically learn a coefficient  $\alpha$ to control the combination between the two parts as:
\begin{equation}\label{eq-copy-prob}
\alpha = \sigma (\bm{w}^\top_{gen}[\tilde{\bm{c}}_{j,t}; \bm{s}_{j,t}]+ b_{gen}).
\end{equation} 

\ignore{To utilize the knowledge information in the generation process, we use the learned state vector $\tilde{\bm{q}}_t$ (Eq.~\ref{eq-att-gru}) as a query to adaptively choose and copy their neighboring entities or words from the graph into the sentence through the copy mechanism. We compute the probability of copying from the graph using global attention vector $\tilde{\bm{q}}_t$ in a way similar to~\cite{SeeLM17} via:
	
	\begin{equation}\label{eq-copy-prob}
	p = \sigma (\bm{W}_{copy}[\bm{h}_{j,t}^Y;\tilde{\bm{q}}_t] + b_{copy}).
	\end{equation} 
	The final word probability distribution is:
	
	\begin{equation}\label{eq-sentence-prob}\small
	\text{Pr}(w_{j,t}|w_{j,<t}, a_j, c) = p \ast \alpha^{copy} + (1-p) \ast \alpha^{vocab},
	\end{equation} 
	where the probability distribution $\alpha^{copy}$ over entities and words in the graph is computed via:
	
	\begin{small}
		\begin{eqnarray}\label{eq-copy-dist}
		\alpha^{copy} &=& \text{softmax} (\bm{W}_4 \tilde{\bm{h}}_{j,t}^Y + \bm{b}_1), \\
		\tilde{\bm{h}}_{j,t}^Y &=& \tanh(\bm{W}_3 [\bm{h}_{j,t}^Y; \tilde{\bm{q}}_t]; \bm{v}_n]),
		\end{eqnarray} 
	\end{small}
}

Here, we present a KG-based copy mechanism. The key point is that we have learned heterogeneous node embeddings  using Caps-GNN. By only copying reachable nodes to the generated words, we hypothesize that there exist semantic dependencies between entities and words in a sentence. Using such a copy mechanism, we can improve the coherence of the sentence content. 
Besides, nodes in the HKG are keywords, entities or items, which makes the generated content more informative.

\subsection{Parameter Learning}
In this part, we discuss the training and inference algorithms for our model. 

Our software environment is built upon ubuntu 16.04, Pytorch v1.1 and python 3.6.2. All the experiments are conducted on a server machine with four GPUs, one CPU and 128G memory.
To learn the model parameters, we factorize the original objective function in Eq.~\ref{eq-joint-prob} into two parts, namely aspect generation and sentence generation. Our parameters, organized by these two parts, are denoted by $\Theta^{(1)}$ and $\Theta^{(2)}$, respectively. 
Algorithm~\ref{alg1} presents the training algorithm for our proposed model. 
The optimization of $\Theta^{(2)}$ for the RNN component is straightforward. The difficulty lies in the learning of $\Theta^{(1)}$, which are parameters of the Caps-GNN.

The loss for aspect generation can be computed through the margin loss defined in Eq.\ref{eq-aspect-loss}. The loss for sentence generation can be computed by summing the negative likelihood of individual words using Eq.\ref{eq-sentence-prob}. 
The joint objective function is difficult to be directly optimized. Hence, we incrementally train the two parts, and fine-tune the shared or dependent
parameters in different modules with the joint objective. For training, we directly use the real aspects and sentences to optimize the model parameters. 
For inference, we apply our model in a pipeline way: we first infer the aspect sequence,  then predict the sentences using inferred aspects.  
During inference, we apply the beam search method with a beam size of 8.
In the aspect and sentence generation modules of our model, we incorporate two special symbols to indicate the start and end of a sequence, namely \textsc{Start} and \textsc{End}. Once we generate the \textsc{End} symbol, the generation process will be stopped.
We set the maximum generation lengths for aspect sequence and review sequence to be 10 and 50, respectively. 
In order to avoid overfitting, we adopt a dropout ratio of 0.2.

The main time cost for our proposed model lies in the capsule graph neural network (Caps-GNN). Since the learning of primary capsules relies on a general R-GCN algorithm, we only focus on the learning of graph capsules and aspect capsules.
In the learning stage of graph capsules, we adopt a dynamic routing mechanism for $\tau$ iterations over $N$ primary capsules and generate $Z$ graph capsules. So the learning of graph capsules achieves $\mathcal{O}(\tau \cdot N \cdot Z)$ time complexity. For efficiency, we only extract a small subgraph from HKG, including items, entities and keywords related to a user. We start with the current user $u$ as the seed, then include its one-hop items and their linked entities, and finally incorporate the keywords.  On average, we can obtain a subgraph with $N_u \ll N$ nodes. So the learning of graph capsules has an average time complexity of $\mathcal{O}(\tau \cdot N_u \cdot Z)$. Note that the graph capsules (Section 4.1) will be learned only once for the HKG in an offline way.
In the learning stage of aspect capsules, we generate adaptive aspect capsules integrating the hidden vector at each time step through a dynamic routing mechanism for $\tau$ iterations. Hence, generating adaptive aspect capsules can be done within $\mathcal{O}(m \cdot \tau \cdot Z \cdot A)$ time complexity, where $m$ is the maximum length of aspect sequence and $A$ is the number of aspect capsules.
Finally, the overall training complexity of our proposed model is $\mathcal{O}(\tau \cdot N_u \cdot Z + m \cdot \tau \cdot Z \cdot A)$.

\ignore{During inference, for sequence generation, we apply the beam search method with a beam size of 4.
In the two sequence generation modules of our model, we incorporate two special symbols to indicate the start and end of a sequence, namely \textsc{Start} and \textsc{End}. Once we generate the \textsc{End} symbol, the generation process will be stopped.
We set the maximum generation lengths for aspect sequence and review sequence to be 10 and 50, respectively. 
In order to avoid overfitting, we adopt the dropout strategy with a rate of 0.2. 
More implementation details can be found in Table~\ref{tab-parameters}.
}

\begin{algorithm}[t]\small
	\caption{The training algorithm for our proposed model.}
	\label{alg1}
	\begin{algorithmic}[1]
		\Require heterogeneous knowledge graph $\mathcal{G}$, learning rate of aspect decoder $\eta^{(1)}$, learning rate of sentence decoder $\eta^{(2)}$
		\State \textbf{Input:} A review dataset $\mathcal{D}$
		\State \textbf{Output:} Model parameters $\Theta^{(1)}$ and $\Theta^{(2)}$
		\State Randomly initialize $\Theta^{(1)}$ and $\Theta^{(2)}$
		\While {not convergence}
		\For{\emph{iteration} $=1$ to $\mathcal{|D|}$}
		\State Acquire $a^{1:m}$ and $w^{1:m}$ for a random context $c=\{u,i,s\}$
		\State $g^{(1)} \leftarrow 0$, $g^{(2)} \leftarrow 0$
		\State Calculate primary capsules according to Eq.~(2)
		\State Calculate graph capsules $\bm{P}$ according to Eq.~(3-6)
		\For{\emph{j} $=1$ to $m$}
		\State Obtain adaptive graph capsules $\bm{\widetilde{P}}$ according to Eq.~(9)
		\State Predict the aspect label $a_j$ according to Eq.~(10)
		\State Calculate margin loss for aspect decoder according to Eq.~(11)
		\State Calculate gradients $\nabla^{(1)}$ for aspect encoder
		\State $g^{(1)} \leftarrow g^{(1)} + \nabla^{(1)}$
		\EndFor 
		\For{$a_1$ to $a_m$}
		\State Calculate the word probability according to Eq.~(14-16)
		\State Calculate the cross-entropy loss for sentence decoder
		\State Calculate gradients $\nabla^{(2)}$ for sentence decoder
		\State $g^{(2)} \leftarrow g^{(2)} + \nabla^{(2)}$
		\EndFor
		\State $\Theta^{(1)} \leftarrow \Theta^{(1)} - \eta^{(1)} * g^{(1)}$, $\Theta^{(2)} \leftarrow \Theta^{(2)} - \eta^{(2)} * g^{(2)}$
		\EndFor
		\EndWhile

		\State \Return $\Theta^{(1)}$ and $\Theta^{(2)}$
	\end{algorithmic}
\end{algorithm}

\section{experiment}
In this section, we first set up the experiments, and then report the results and analysis.

\subsection{Experimental Setup}

\subsubsection{Construction of the Datasets} To measure the performance of our proposed model, We use three real-world datasets from different domains,  including \textsc{Amazon} Electronic~\cite{HeM16}, Book datasets~\cite{HeM16}, and \textsc{IMDb} Movie dataset\footnote{https://www.imdb.com}. 
In order to obtain KG information for these items, we 
adopt the public KB4Rec~\cite{zhao2019kb4rec,HuangZDWC18} dataset and follow its method to construct the aligned linkage between Freebase~\cite{freebase} (March 2015 version) entities and online items from the three domains. 
All the text is processed with the procedures of lowercase, tokenization, and infrequent word removal (only keeping top frequent 30,000 words). 
We also remove users and products (or items) occurring fewer than five times, and discard reviews containing more than 100 tokens. Note that not all the items can be aligned to Freebase entities, and we only keep the data of the aligned items.
Starting with the aligned items as seeds, we include their one-hop neighbors from Freebase as our KG data.
We removed relations like \emph{<book.author.written\_book>} which just reverses the head and tail compared to the relations \emph{<book.written\_book.author>}. We also remove relations that end up with non-freebase string, \eg like \emph{<film.film.rottentomatoes\_id>}.
We summarize the statistics of three datasets after preprocessing in Table~\ref{tab-data}. 
Furthermore, for each domain, we randomly split it into training, validation  and test sets with a ratio of 8:1:1.


\begin{table}[htbp]
	\centering
	\caption{Statistics of our datasets after preprocessing.}
	\begin{tabular}{|c|l|r|r|r|}
		\hline
		\multicolumn{2}{|c|}{Dataset}& Electronic & Book & Movie\\
		\hline
		\hline
		\multirow{3}{*}{\tabincell{c}{Review}}&\#Users 	&	50,473&	71,156&	47,096\\
		&\#Items 	&12,352&	25,045&	21,125\\
		&\#Reviews 	&	221,722&	853,427&	1,152,925\\
		\hline
		\multirow{3}{*}{\tabincell{c}{Knowledge\\Graph}}&\#Entities	&30,310	&105,834	&247,126	\\
		&\#Relations	&15	&10	&13	\\
		&\#Triplets 	&129,254	&300,416	&1,405,348	\\
		\hline
	\end{tabular}%
	\label{tab-data}%
\end{table}%

\subsubsection{Aspect and Opinion Extraction} 
To construct our HKG, we need to incorporate word nodes. We only consider aspect and opinion keywords, which are more important  for review text. 
We use the Twitter-LDA model in~\cite{ZhaoJWHLYL11} for automatically learning the aspects and aspect keywords. The numbers of aspects are all set to 10 for the three datasets.  
With topic models, we  tag each sentence with the aspect label which gives the maximum posterior probability conditioned on the keywords. 
For each domain, we keep the words ranked in top 70 positions of each aspect as aspect keywords.
After obtaining the aspect keywords, we leverage four syntactic rules in~\cite{QiuLBC11} (\eg ``OP~(\textbf{JJR}) $\stackrel{\text{amod}}{\longrightarrow}$ AP~(\textbf{NN})" ) to identify the potential opinion keywords. For example, the rule ``OP~(\textbf{JJR}) $\stackrel{\text{amod}}{\longrightarrow}$ AP~(\textbf{NN})" means that the opinion keywords (OP) often occur ahead of aspect keywords (AP).
We keep the top 200 opinion keywords in the entire text collection, such as ``charming'' and ``perfect''.  
To identify entity mentions, we employ a strict string match and filter ambiguous candidates using the semantics of the current item.
Although this method tends to miss entity mentions, it can achieve a high precision and provide sufficient information to construct the entity-word links. For reducing noise, we only keep the top 50\%  keywords that co-occur with an entity for link creation. 

\subsubsection{Baseline Methods} We compare our model against the following methods:

\textbullet~\textmd{gC2S}~\cite{TangYCZM16}: It applies an encoder-decoder framework to generate review texts conditioned on context information through a gating mechanism.

\textbullet~\emph{Attr2Seq}~\citep{ZhouYHZXZ18}: It adopts an attention-enhanced attribute to sequence architecture to generate reviews with input attributes (\eg user, item and rating).

\textbullet~\emph{Attr2Seq+KG}: We incorporate the pre-trained KG embeddings of items as additional inputs into Attr2Seq.



\textbullet~\textmd{SeqGAN}~\cite{YuZWY17}: It regards the generative model as a stochastic parameterized policy and uses Monte Carlo search to approximate the state-action value. The discriminator is a binary classifier to evaluate the sequence and guide learning process of the generator.

\textbullet~\emph{LeakGAN}~\citep{GuoLCZYW18}: It is designed for long text generation through the leaked mechanism.
The generator is built upon a hierarchical reinforcement learning architecture and the discriminator is a CNN-based feature extractor. 

\textbullet~\emph{ExpansionNet}~\citep{NiM18}: It builds an encoder-decoder architecture to generate personalized reviews by introducing aspect-level information (\eg aspect words) and short phrases (\eg review summaries, product titles). 

\textbullet~\emph{AP-Ref2Seq}~\citep{NiLM19}: It employs a reference-based Seq2Seq model with aspect-planning which can generate personalized reviews covering different aspects.

\textbullet~\emph{ACF}~\citep{LiZWS19}: It decomposes the review generation process into three different stages by designing an aspect-aware coarse-to-fine generation model. The aspect semantics and syntactic characteristics are considered in the process.

Among these baselines, gC2S and Attr2Seq are context-aware generation models in different implementation approaches; SeqGAN and LeakGAN are GAN-based text generation models; ExpansionNet, AP-Ref2Seq and ACF incorporate external aspect information as input;  ACF is the state-of-the-art review generation model. 
Additionally, to examine the usefulness of KG incorporation, we build an Attr2Seq+KG model by integrating pre-trained item KG embeddings into Attr2Seq as additional attribute input. We use DistMult~\cite{YangYHGD14a} to pre-train KG embeddings.
We employ validation set to optimize the parameters and select the optimal parameters in each method. 
To reproduce the results of our model, we report the parameter setting used throughout the experiments in Table~\ref{tab-parameters}. 
	
\begin{table}[htbp]
	\centering
	\caption{Parameter settings of the two modules in our model.}\label{tab:parameters}
	\begin{tabular}{|c | c|}
		\hline
		Modules & Settings\\\hline\hline
		Aspect & \tabincell{l}{$d_E=512$, $d_H=512$, $d_C=100$, \\
			$Z$=10, $d_A=512$, batch-size=$1024$,\\
			\#GCN-layer=$3$, \#GRU-layer=$2$,  \\ 
			init.-learning-rate=$0.00002$, Adam optimizer}\\
		\hline
		Review & \tabincell{l}{$d_W=512$, $d_S=512$, \\ \#GRU-layer=$2$, 
			batch-size=$64$,\\ init.-learning-rate=$0.0002$, \\ learning-rate-decay-factor=0.8, \\
			learning-rate-decay-epoch=2, Adam optimizer}\\
		\hline
	\end{tabular}
	\label{tab-parameters}
\end{table}

\subsubsection{Evaluation Metrics} To evaluate the performance of different methods on automatic review generation, 
we  adopt six evaluation metrics, including Perplexity, BLEU-1/BLEU-4, ROUGE-1/ROUGE-2/ROUGE-L. Perplexity\footnote{https://en.wikipedia.org/wiki/Perplexity} is a standard measure for evaluating language models; BLEU~\cite{PapineniRWZ02} measures the ratio of the co-occurrence of $n$-grams between the generated and real reviews;
 and ROUGE~\cite{Lin04} measures the review quality by counting the overlapping $n$-grams between the generated and real reviews.

\subsection{Performance Comparison}

We present the results of different methods on the review generation task in Table~\ref{tab:main-results}.

First, among the two simple methods (namely gC2S and Attr2Seq), it seems that Attr2Seq is slightly better than gC2S. The difference between Attr2Seq and gC2S is that Attr2Seq utilizes the attention mechanism to incorporate attribute information, while gC2S utilizes a simpler gate mechanism. 
Furthermore, by incorporating the KG embeddings, Attr2Seq+KG achieves  better results than Attr2Seq, which indicates the effectiveness of KG data.

Second, there exists an inconsistent trend for GAN-based methods. It seems that LeakGAN performs better than the above simple methods, while  SeqGAN seems to give worse results.  A major reason is that LeakGAN is specially designed for generating long text, while the rest GAN-based methods may not be effective in capturing long-range semantic dependency in text generation. 

Third, by incorporating aspect words and other attribute information,  ExpansionNet, AP-Ref2Seq and ACF perform better than Attr2Seq and its KG-enhanced version Attr2Seq+KG. It shows that aspect information is helpful for review generation and simply incorporating KG information cannot yield very good performance.
The most recently proposed method ACF performs best among all the baselines. 
It  adopts a three-stage generation process by considering both aspect semantics and syntactic patterns. 



Finally, our model outperforms all the baselines with a large margin. 
The major difference between our model and ACF lies in that KG information has been utilized in the multi-stage generation process. 
ACF fully relies on sequential neural networks to learn from training text, while we use KG data to instruct the generation of aspect sequences and sentence sequences. 
In particular, we utilize Caps-GNN and copy mechanism to capture the user preference at both aspect and word levels, which yields a better performance than all baselines. 


\subsection{Detailed Analysis}
In this part, we construct a series of experiments on the effectiveness of the proposed model.
We will only report the results on \textsc{Movie} dataset due to similar findings in three datasets. We select the  three best baselines \emph{LeakGAN},  \emph{ExpansionNet} and \emph{ACF} as comparisons.

\subsubsection{Ablation Analysis} Based on previous review generation studies~\cite{NiM18, LiZWS19}, our model has made several important extensions. First, we construct a HKG as additional data signal to improve the PRG task. Second, we propose a novel capsule GNN for capturing aspect semantics. Third, we utilize copy mechanism to generate important entity words. Here, we would like to examine how each factor  contributes to the final performance.
To see this, we prepare five variants for comparison:


$\bullet$ \emph{w/o KG}: the variant removes the KG entities and their links from HKG, but keep the other nodes and links. 

$\bullet$ \emph{w/o HKG, w KG}: the variant removes the user and word nodes from HKG but retains the KG entities and their links.

$\bullet$ \emph{w/o Caps-GNN, w R-GCN}: the variant replaces the Caps-GNN with a conventional R-GCN component. 

$\bullet$ \emph{w/o Caps-GNN, w GAT}: the variant replaces the Caps-GNN with a conventional GAT component.

$\bullet$ \emph{w/o Copy}: the variant removes the copy mechanism during generating reviews. 

Table~\ref{tab:ablation-results} presents the performance comparison between the complete model and the five variants.  
We can see that removing KG data significantly affects the performance of our model, which further verifies the usefulness of KG data. 
Besides, the variant removing user and word nodes gives a worse result than the complete model, which shows that HKG is better for our task than KG. 
Third, variants dropping the Caps-GNN component are worse than the complete model, which indicates Caps-GNN is better to capture user preference than other GNN methods.
Finally, removing the copy mechanism greatly declines the performance of our model. 
In our model, the copy mechanism directly generates word tokens by selecting 
related  entities or words from HKG, which has a more significant effect on the final performance. 
This observation also implies that  real reviews indeed contain important entity information, and the generation model should incorporate KG data for a better performance. 

\begin{figure}[t]
	\centering
	\subfigure[Varying the amount of KG triples.]{\label{fig-varing-kg}
		\centering
		\includegraphics[width=0.215\textwidth]{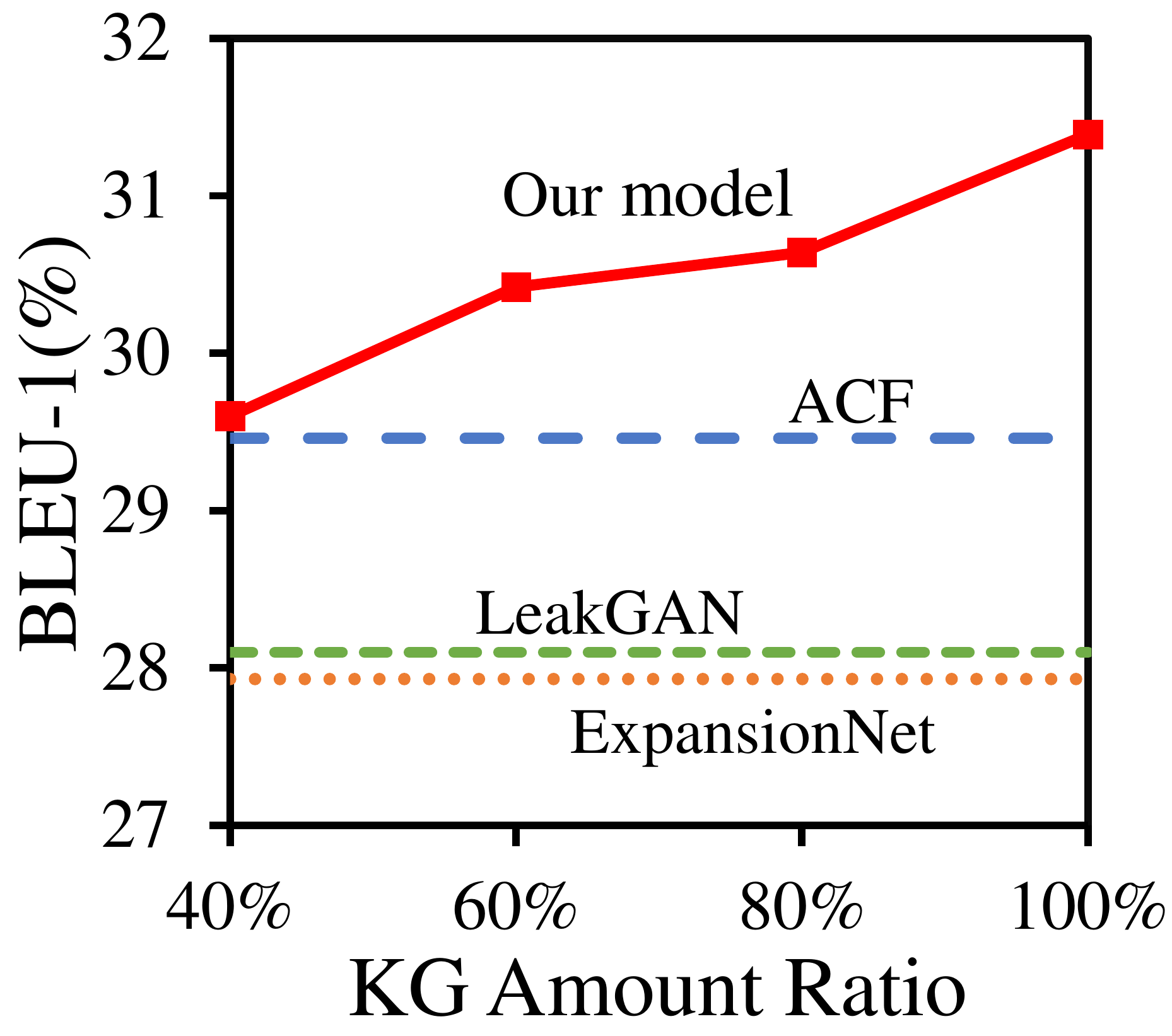}
	}
	\subfigure[Varying the embedding size.]{\label{fig-varing-es}
		\centering
		\includegraphics[width=0.21\textwidth]{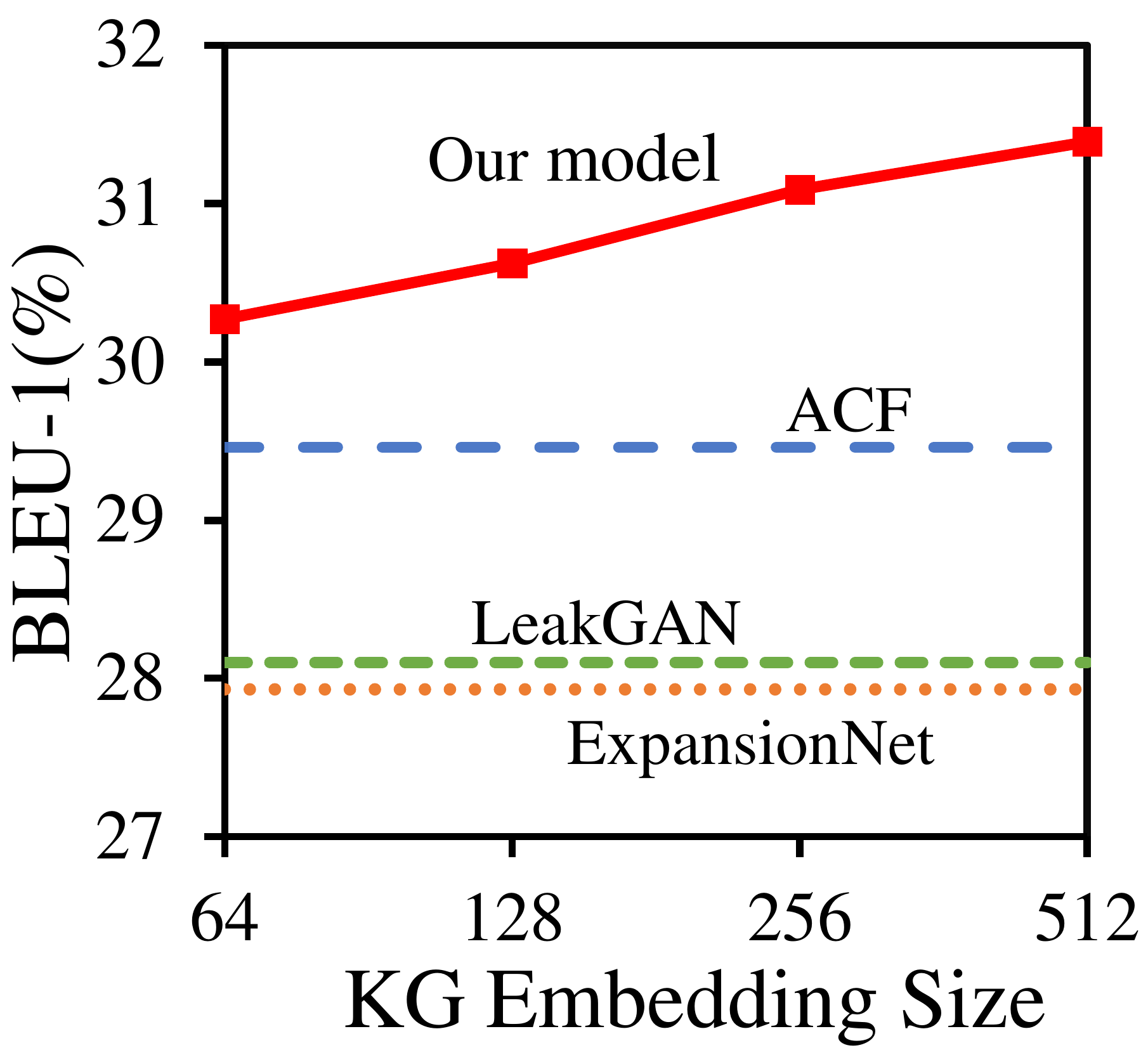}
	}
	\centering
	\caption{Performance tuning  on \textsc{Movie} dataset.}
	\label{fig-parameter-tuning}
\end{figure}

\begin{table*}[tb]
	\renewcommand\arraystretch{1.1}
	\begin{center}
		\caption{Performance comparisons of different methods for automatic review generation under three domains. ``*'' denotes the improvement is stastically significant compared with the best baseline (t-test with p-value $< 0.05$).}
		\begin{tabular}{|c||l||c c c c c c|}
			\hline
			\textmd{Datasets} & \textmd{Models} & \textmd{Perplexity} & \textmd{BLEU-1(\%)} & \textmd{BLEU-4(\%)} & \textmd{ROUGE-1} & \textmd{ROUGE-2} & \textmd{ROUGE-L} \\
			\hline \hline
			\multirow{4}[6]{*}{\textsc{Electronic}}
			& gC2S & 38.67 & 24.14 & 0.85 & 0.262 & 0.046 & 0.212 \\
			& Attr2Seq & 34.67 & 24.28 & 0.88 & 0.263 & 0.043 & 0.214 \\
			& Attr2Seq+KG & 33.12 & 25.62 & 0.93 & 0.271 & 0.049 & 0.223 \\
			\cline{2-8}
			& SeqGAN & 28.50 & 25.18 & 0.84 & 0.265 & 0.043 & 0.220 \\
			& LeakGAN & 27.66 & 25.66 & 0.92 & 0.267 & 0.050 & 0.236 \\
			\cline{2-8}
			& ExpansionNet & 31.50 & 26.56 & 0.95 & 0.290 & 0.052 & 0.262 \\
			& AP-Ref2Seq & 27.59 & 27.04 & \underline{1.15} & 0.309 & 0.065 & 0.279 \\
			& ACF & \underline{26.55} & \underline{28.22} & 1.04 & \underline{0.315} & \underline{0.066} & \underline{0.280} \\
			\cline{2-8}
			& Our model & \textbf{26.05} & \textbf{29.88*} & \textbf{1.83} & \textbf{0.323*} & \textbf{0.078*} & \textbf{0.295*} \\
			\hline \hline
			\multirow{4}[6]{*}{\textsc{Book}}
			& gC2S & 30.58 & 25.87 & 1.03 & 0.265 & 0.044 & 0.217 \\
			& Attr2Seq & 30.87 & 26.93 & 1.14 & 0.259 & 0.047 & 0.223 \\
			& Attr2Seq+KG & 30.33 & 27.69 & 1.42 & 0.268 & 0.053 & 0.236 \\
			\cline{2-8}
			& SeqGAN & 27.11 & 26.89 & 1.24 & 0.255 & 0.053 & 0.246 \\
			& LeakGAN & 25.79 & 28.79 & 1.94 & 0.274 & 0.060 & 0.285 \\
			\cline{2-8}
			& ExpansionNet & 28.76 & 26.52 & 1.49 & 0.301 & 0.054 & 0.271 \\
			& AP-Ref2Seq & 25.38 & 28.34 & 1.82 & \underline{0.318} & \underline{0.075} & 0.283 \\
			& ACF & \underline{24.38} & \underline{28.96} & \underline{2.11} & 0.317 & 0.068 & \underline{0.291}  \\
			\cline{2-8}
			& Our model & \textbf{22.24*} & \textbf{30.66*} & \textbf{3.08*} & \textbf{0.332*} & \textbf{0.080*} & \textbf{0.306*} \\
			\hline \hline
			\multirow{4}[6]{*}{\textsc{Movie}}
			& gC2S & 34.12 & 26.17 & 1.09 & 0.272 & 0.047 & 0.215 \\
			& Attr2Seq & 33.12 & 26.57 & 1.55 & 0.271 & 0.050 & 0.222 \\
			& Attr2Seq+KG & 34.19 & 27.02 & 1.67 & 0.278 & 0.053 & 0.235 \\
			\cline{2-8}
			& SeqGAN & 24.53 & 27.07 & 1.63 & 0.274 & 0.052 & 0.221 \\
			& LeakGAN & \textbf{21.76} & 28.10 & 2.29 & 0.302 & 0.064 & 0.271 \\
			\cline{2-8}
			& ExpansionNet & 27.94 & 27.93 & 2.00 & 0.310 & 0.063 & 0.266 \\
			& AP-Ref2Seq & 24.78 & 29.01 & 2.12 & 0.314 & 0.074 & \underline{0.306} \\
			& ACF & \underline{22.68} & \underline{29.46} & \underline{2.40} & \underline{0.322} & \underline{0.076} & 0.303  \\
			\cline{2-8}
			& Our model & 23.34 & \textbf{31.39*} & \textbf{3.55*} & \textbf{0.341*} & \textbf{0.096*} & \textbf{0.327*} \\
			\hline
		\end{tabular}
		\label{tab:main-results}
	\end{center}
\end{table*}

\begin{table}[t]
	\centering
	\caption{Ablation analysis on \textsc{Movie} dataset.} 
	\begin{tabular}{| l || c c  |}
		\hline
		Models & BLEU-1(\%) & ROUGE-1 \\
		\hline
		\hline
		Complete model & 31.39 & 0.341 \\
		\hline
		w/o KG & 29.19 & 0.320 \\
		w/o HKG, w KG & 30.56 & 0.335 \\
		w/o Caps-GNN, w R-GCN& 30.45 & 0.333 \\
		w/o Caps-GNN, w GAT& 29.79 & 0.331 \\
		w/o Copy & 29.02 & 0.322 \\
		\hline
	\end{tabular}
	\label{tab:ablation-results}
\end{table}

\begin{table}[t]
	\centering
	\caption{Aspect coverage evaluation on \textsc{Movie} dataset.} 
	\begin{tabular}{| l || l  l  l |}
		\hline
		Models & \tabincell{l}{\# aspects\\(real)} & \tabincell{l}{\# aspects\\(generated)} & \tabincell{l}{\# covered\\aspects} \\
		\hline
		\hline
		LeakGAN & 4.16 & 2.82 & 1.039 \\
		ExpansionNet & 4.16 & 2.94 & 1.829 \\
		ACF & 4.16 & \uline{3.11} & \uline{2.105} \\
		Our model & 4.16 & \textbf{3.25} & \textbf{2.853} \\
		\hline
	\end{tabular}
	\label{tab:aspect-results}
\end{table}

\begin{table}[tb]
	\centering
	\caption{Human evaluation on  three dimensions. ``Gold'' indicates the ground-truth reviews.}
	\begin{tabular}{|l || c  c  c c |}
		\hline
		Models & Relevance &  Informativeness & Fluency &\\
		\hline
		\hline
		Gold & 4.25 & 3.93 & 4.33 &\\
		\hline
		LeakGAN & 3.50 & 2.93 & 3.40 &\\
		ExpansionNet & 3.68 & 2.38 & 3.23 &\\
		ACF & 3.48 & 2.68 & 3.63 &\\
		Our model & 3.95 & 3.53 & 3.50 & \\
		\hline
	\end{tabular} 
	\label{tab:human-results}
\end{table}

\begin{table*}[tb]
	\renewcommand\arraystretch{1.1}
	\begin{center}
		\begin{small}
			\caption{Sample reviews generated by our model from \textsc{IMDb} movie dataset. The two reviews are about  the movies of \emph{``Sleepy Hollow''} and \emph{``Gladiator''} from the same user. 
				Colored, boxed and underlined words correspond to aspect labels (created for reading), entities and keywords. 
				The first column corresponds to the actual inclusions of triples and entity-word pairs by our model in the generated review. 
			}
			\begin{tabular}{c|| c ||c }
				\hline
				\textmd{Heterogeneous Knowledge Graph} & \textmd{Gold Standard} & \textmd{Generated Review}  \\
				\hline \hline
				\begin{minipage}{0.27\textwidth}\includegraphics[width=45mm, height=20mm]{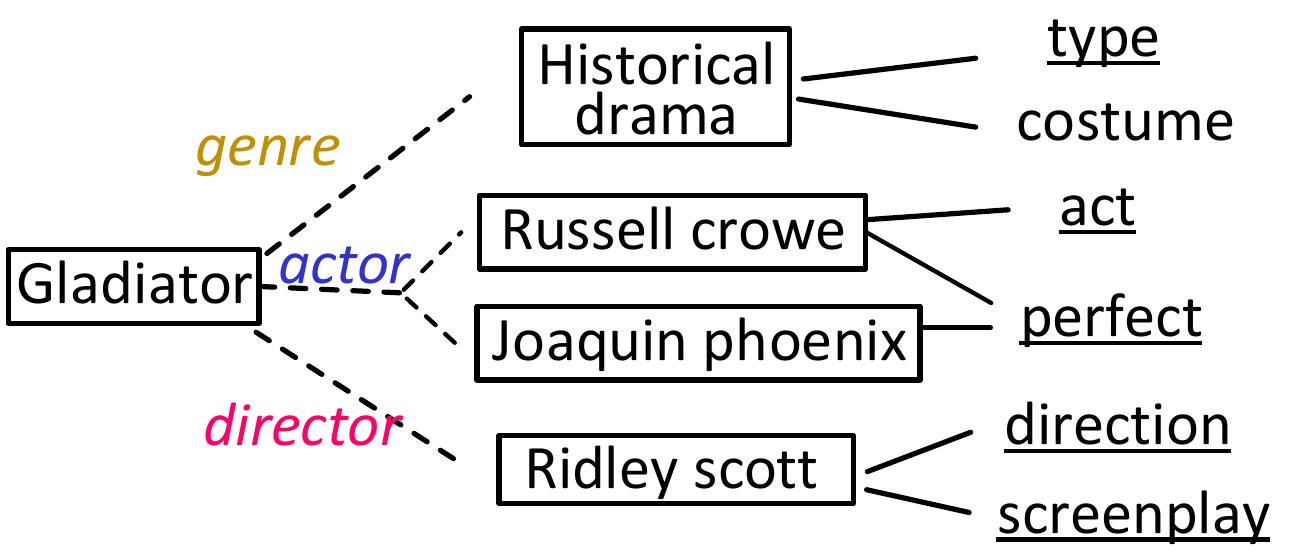}\end{minipage}
				
				& \tabincell{l}{ this movie lives up to \textbf{mr. burton} standards and we \\ \underline{expect} from this brilliant $\text{\underline{director}}_{\textcolor[RGB]{238,48,167}{\text{director}}}$ the story is \\  $\text{good}_{\textcolor[RGB]{34,139,34}{\text{story}}}$ there are a few great \underline{actors} in the $\text{\underline{cast}}_{\textcolor[RGB]{0,0,255}{\text{actor}}}$ \\ and the movie is a \underline{fantasy} \textbf{adventure} romp but it is \\ not one for the $\text{kids}_{\textcolor[RGB]{139,69,19}{\text{genre}}}$ it has a \underline{distinct gothic feel} \\ though not for the squeamish or faint of $\text{heart}_{\textcolor[RGB]{255,0,0}{\text{overall}}}$}
				
				& \tabincell{l}{this is one of the best \underline{fantasy} \textbf{adventure} movies \\i have seen in a long $\text{time}_{\textcolor[RGB]{139,69,19}{\text{genre}}}$ i have to say that \\i am a huge fan of \textbf{tim burton} and \underline{expected} more \\from a \textbf{tim burton} $\text{film}_{\textcolor[RGB]{238,48,167}{\text{director}}}$ the film is full of fun \\with a great \underline{cast} , including a great \underline{performance} \\by $\textbf{johnny depp}_{\textcolor[RGB]{0,0,255}{\text{actor}}}$ it has $\text{\underline{distinct feel}}_{\textcolor[RGB]{255,0,0}{\text{overall}}}$}   \\ 
				
				\hline
				
				\begin{minipage}{0.27\textwidth}\includegraphics[width=45mm, height=21mm]{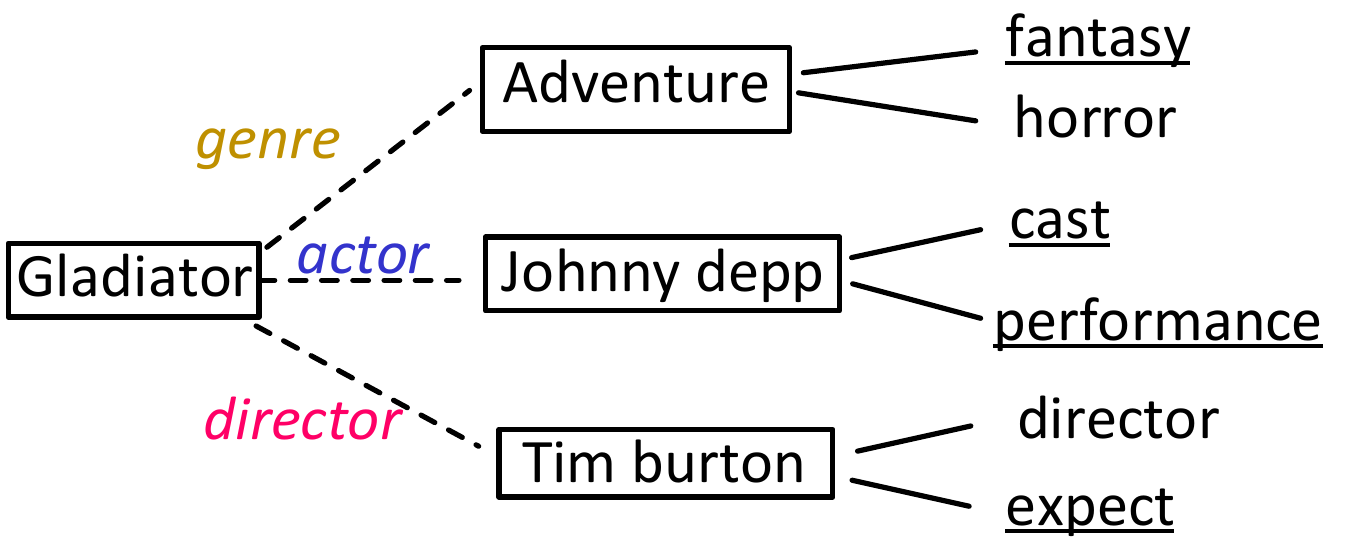}\end{minipage}
				
				& \tabincell{l}{ \textbf{gladiator} is not just \underline{director} \textbf{ridley scott} ultimate\\ $\text{masterpiece}_{\textcolor[RGB]{238,48,167}{\text{director}}}$ it
					had its moments as a costume \\\textbf{drama} and was good to look $\text{at}_{\textcolor[RGB]{139,69,19}{\text{genre}}}$ the music may \\\underline{sound} quite stuffy but fits perfectly for the $\text{film}_{\textcolor[RGB]{34,139,34}{\text{music}}}$ \\the \underline{characters} are all great . \textbf{pheonix} is annoying \\but he \underline{acts} his character \underline{perfectly} , $\text{pretentious}_{\textcolor[RGB]{0,0,255}{\text{actor}}}$ } 
				
				& \tabincell{l}{\textbf{gladiator} is a great movie to see if you like the \\\underline{type} of  $\textbf{historical drama}_{\textcolor[RGB]{139,69,19}{\text{genre}}}$ the \underline{acting} is excel-\\lent and \underline{perfect} especially from \textbf{russell crowe} and\\ $\textbf{joaquin phoenix}_{\textcolor[RGB]{0,0,255}{\text{actor}}}$ i was surprised by great \\\underline{direction} and \underline{screenplay} by $\textbf{ridley scott}_{\textcolor[RGB]{238,48,167}{\text{director}}}$ it is \\a must see for any fan and you will love $\text{it}_{\textcolor[RGB]{255,0,0}{\text{overall}}}$ }  \\
				\hline
			\end{tabular}
			\label{tab:examples}
		\end{small}
	\end{center}
\end{table*}

\subsubsection{Aspect Coverage Evaluation} 
A major motivation of our work is to improve the generation of informative words via aspect modeling. Following~\cite{NiM18}, we perform the evaluation by measuring how many aspects in real reviews are covered in generated reviews.
Since we have obtained topic models for all the aspects, we consider a (ground-truth or generated) review as covering an aspect if any of the top 50 keywords  of an aspect exists in the review. For guaranteeing the quality of  topic words,  we manually remove irrelevant or noisy words from the top 50 keywords. 

We present the aspect coverage results of different methods in Table~\ref{tab:aspect-results}.  First, we can see that LeakGAN and ExpansionNet have generated similar numbers of aspects (2.82 \emph{vs} 2.94), while ExpansionNet has covered  a more significant number of real aspects than LeakGAN (1.829 \emph{vs} 1.039).
LeakGAN is not tailored to the review generation task, while ExpansionNet incorporates aspect information into generation model. Then,  ACF performs best among the three baselines. It also sets up an  aspect generation component based on GRU decoder and context information. 
Finally, it can observed that our model is able to generate more aspects and cover more real aspects.
Compared with ACF, our model has a similar number of generated aspects, but a larger number of covered aspects. 
The reason lies in that our model can capture aspect-level user preference from the HKG and generate personalized aspect sequence through Caps-GNN.

\ignore{ see that LeakGAN and ExpansionNet are worse than the other baselines varying the generated aspects and covered aspects by a large margin. 
	As a comparison, ACF and our model perform better than LeakGAN and ExpansionNet by generating more aspects, since the two models characterize the aspect transition sequence before generating review sentences. However, due to the utilization of knowledge graph, our model covers more real aspects than Coarse-to-Fine. These results indicate the usefulness of knowledge graph in capturing more personalized information related to a user.}

\subsubsection{Model Sensitivity \emph{w.r.t.} KG Data} 
In previous experiments, we have shown that KG data is indeed very helpful to improve the 
performance of our model. Here, we would examine how it affects the final performance.
In this part, we fix the review data (including both training and test) as original, and vary the part for KG data. For comparisons, we take the best performance results of LeakGAN, ExpansionNet and ACF as references. 

We first examine the effect of  the amount of KG data on the performance. We gradually increase the available data for training from   40\% to 100\% with a step of 20\%. 
In this way, we generate four new KG datasets. 
We utilize them together with the original review data to train our model, and evaluate on the test set.  As shown in Fig.~\ref{fig-parameter-tuning}(a), the performance of our model gradually improves with the increasing of KG data, and our model has achieved a consistent improvement over ACF with more than 40\% KG data.

For KG data, the embedding size is an important parameter to tune in real applications. 
Here, we vary the embedding size in the set $\{ 64, 128, 256, 512\}$. We construct a similar evaluation experiment as that for the amount of KG data. 
In Fig.~\ref{fig-parameter-tuning}(b), we can see that our model is consistently better than the two selected baselines with four sets. The embedding size of 512 yields the best results for our model, while the improvement seems to become small when it is larger than 256.

\ignore{ KG data provides an  important data signal to improve the PRG task for our model. Here, we examine how it affects the final performance. 
First, for the amount of KG data, we take 40\%, 60\%, 80\% and 100\% from the complete training data to generate four new training sets, respectively. Second, for KG embedding size, we select 64, 128, 256 and 512 to train our model. The test set is fixed as original. As shown in Fig.~\ref{fig-parameter-tuning}, our model is consistently better than the three baselines in all cases. The BLEU-1 performance improves when having more KG triples, and the KG embedding size of 512 yield the best results for our model. 
}

\subsection{Human Evaluation} 
Above, we have performed automatic evaluation experiments for our model and baselines. 
For text generation models, it is important to construct human evaluation for further effectiveness verification.

We randomly select 200 sample reviews from the test set of the Movie dataset. 
A sample review contains the input information (including user, item and rating) and its ground-truth review. Given a sample review, we collect the generated reviews from different models, and then shuffle them for human evaluation.  
Following~\cite{KielaWZDUS18}, we invite two human judges to read all the results and assign scores to a generated review with respect to three factors of quality, relevance, informativeness, and fluency.
According to \cite{KielaWZDUS18},
\emph{relevance} means that how relevant the generated text is according to the input contexts,
\emph{informativeness} means that how much the generated text provides specific or different  information,
and \emph{fluency} means that how likely the generated text is produced by human.

We adopt a 5-point Likert scale~\cite{likert1932technique} as the  scoring mechanism, in which 5-point means ``very satisfying'', and 1-point means ``very terrible''~\cite{likert1932technique}. 
For each method, we average the  scores from the two human judges and then report the average results. 
We present the results of human evaluation in  Table \ref{tab:human-results}.
It can be seen that our model is  better than the two baselines with a  large margin in terms of \emph{relevance} and \emph{informativeness}. The major reason is that 
we utilize KG data to effectively enrich the generated text with more informative content.   
The fluency of our model is slightly worse than ACF. It is possibly because ACF has considered more syntactic patterns, such as part-of-speech tags and n-grams.
Indeed, it is straightforward to incorporate such linguistic features  to improve the fluency of  our model. 
While, it is not our focus in this work, and will leave it as future work. 
The Cohen’s kappa coefficients are 0.76 in \emph{relevance}, 0.72 in \emph{informativeness} and 0.74 in \emph{fluency}, indicating a high correlation and agreement between the two human judges.

\subsection{Qualitative Analysis}
Previous experiments have demonstrated the effectiveness of our model in generating high-quality review text. Here, we further present intuitive explanations
why our model can generate review texts reflecting user preference through qualitative analysis.

Table~\ref{tab:examples} presents two \textsc{IMDb} movie reviews and the corresponding generated texts by our model. The two reviews are written by the same user about two different movies. By reading the ground-truth reviews, we can infer that  the user mainly focuses on three major aspects, namely \emph{genre}, \emph{actor} and \emph{director}. 
It indicates that users are likely to have an aspect-level preference when writing the review text.
As we can see, our model can capture aspect-level preference and cover most of real aspects, which is helpful to generate personalized sentences.  Interestingly, the involved KG relations have aligned well with real aspects. This implies that the KG data can provide important aspect semantics for learning user preference.

 Furthermore,  the generated text  is highly informative, containing 
 important and personalized entities related to the user about two movies. 
 For example, through capturing aspect-level user preference, the directors and genres that user prefers (\eg \emph{Tim Burton}, \emph{adventure}) have been generated by the KG-enhanced copy mechanism in our model. It is  difficult to directly predict such entities using simple RNN-based text generation model. 
 KG information can be considered as an important data signal to enhance the text generation capacity.
By associating entities with modifier words in the constructed HKG, our model can produce clear, fluent text segments, \eg ``the best \emph{fantasy}$_{modifier}$ \underline{adventure}$_{entity}$ movies''.
 
The above qualitative example has shown that our model can generate personalized review texts 
with both \emph{aspect-} and \emph{word-level} semantics by incorporating KG data. 
 



\section{Conclusion}
In this paper, we have developed a novel KG-enhanced review generation model for automatically generating informative and personalized review text. 
Our core idea is to utilize structural KG data to improve the generated text by incorporating both aspect- and word-level semantics. 
For this purpose, we constructed a HKG by augmenting the original KG with user and word nodes. By constructing a HKG, we can learn graph capsules using a Caps-GNN for capturing underlying KG semantics from different aspects. 
We designed an aspect-aware two-stage text generation model. In this model, we learned adaptive aspect capsules based on  graph capsules to instruct the prediction for the aspect label. Furthermore, we designed a KG-based copy mechanism for directly incorporating related entities or words from KG. 
We constructed extensive experiments on three real-world review datasets. The results showed that our proposed model is superior to previous methods in a series of evaluation metrics for the PRG task. 

Currently, only three datasets with aligned entity-item linkage have been used for evaluation. We believe our approach is applicable to more domains. As future work, we will consider integrating more kinds of external knowledge (\eg WordNet) for the PRG task.

\section*{Acknowledgement}
This work was partially supported by the National Natural Science Foundation of China under Grant No. 61872369 and 61832017, the Fundamental Research Funds for the Central Universities, the Research Funds of Renmin University of China under Grant No.18XNL G22 and 19XNQ047, Beijing Academy of Artificial Intelligence (BAAI) under Grant No. BAAI2020ZJ0301, and Beijing Outstanding Young Scientist Program under Grant No. BJJWZYJH012019100020098. Xin Zhao is the corresponding author.

\bibliographystyle{ACM-Reference-Format}
\bibliography{review_bib}

\end{document}